\documentclass[conference,9pt]{IEEEtran}
\IEEEoverridecommandlockouts

\usepackage{cite}
\usepackage[hidelinks]{hyperref}
\usepackage{amsmath,graphicx,cleveref}
\usepackage{graphicx}
\usepackage{multirow}
\usepackage{textcomp}
\usepackage{hhline}
\usepackage{xcolor}
\usepackage{algorithm}
\usepackage{algpseudocode}
\usepackage[acronym]{glossaries}
\DeclareMathOperator*{\argmax}{arg\,max}
\makeatletter
\def\@IEEEfigurecaptionsepspace{\vskip0.35\baselineskip\relax}
\makeatother

\newcommand{\papertitle}{A Systematic Comparison of Search Space Complexity and Efficiency for Time-Synchronous End-to-End ASR Models}
\title{\papertitle}

\newif\ifanonymous
\anonymousfalse
\newcommand{\toolkitname}{\ifanonymous \textit{ToolkitName}\else RASR\fi}

\newif\ifarxiv
\arxivtrue
\ifarxiv
\usepackage{eso-pic}
\AddToShipoutPictureBG*{%
    \AtPageUpperLeft{%
        \raisebox{-0.4in}{\makebox[\paperwidth]{\parbox{0.8\paperwidth}{\centering\scriptsize
            \copyright~2026 IEEE. Personal use of this material is permitted. Permission from IEEE must be obtained for all other uses, in any current or future media, including reprinting/republishing this material for advertising or promotional purposes, creating new collective works, for resale or redistribution to servers or lists, or reuse of any copyrighted component of this work in other works.}}}}}
\fi

\newacronym{ASR}{ASR}{Automatic Speech Recognition}
\newacronym{LM}{LM}{Language Model}
\newacronym{CTC}{CTC}{Connectionist Temporal Classification}
\newacronym{BPE}{BPE}{Byte-Pair Encoding}
\newacronym{ILM}{ILM}{Internal Language Model}
\newacronym{LSTM}{LSTM}{Long Short-Term Memory}
\newacronym{WER}{WER}{Word Error Rate}
\newacronym{HMM}{HMM}{Hidden Markov Model}
\newacronym{LVCSR}{LVCSR}{large vocabulary continuous speech recognition}
\newacronym{WFST}{WFST}{weighted finite state transducer}
\newacronym{E2E}{E2E}{end-to-end}
\newacronym{RNN}{RNN}{Recurrent Neural Network}
\newacronym{AM}{AM}{Acoustic Model}
\newacronym{RTF}{RTF}{Real-Time Factor}
\newacronym{AED}{AED}{Attention Encoder-Decoder}
\newacronym{OCLR}{OCLR}{One-Cycle Learning Rate}
\newacronym{LLM}{LLM}{Large Language Model}
\newacronym{EOW}{EOW}{end-of-word}

\ifanonymous
\author{Anonymous Author(s)}
\else
\author{\IEEEauthorblockN{Simon Berger}
    \IEEEauthorblockA{\textit{Human Language Technology and Pattern Recognition,}\\
    \textit{Computer Science Department, RWTH Aachen University, Germany}\\
    \textit{AppTek GmbH, Germany}\\
    \texttt{berger@hltpr.rwth-aachen.de}\\}
\and
\IEEEauthorblockN{Larissa Kleppel}
\IEEEauthorblockA{\textit{Human Language Technology and Pattern Recognition,}\\
\textit{Computer Science Department, RWTH Aachen University, Germany}\\
\texttt{lkleppel@hltpr.rwth-aachen.de}\\}
\and
\IEEEauthorblockN{Ralf Schlüter}
\IEEEauthorblockA{\textit{Human Language Technology and Pattern Recognition,}\\
\textit{Computer Science Department, RWTH Aachen University, Germany}\\
\textit{AppTek GmbH, Germany}\\
\texttt{schlueter@hltpr.rwth-aachen.de}
}}
\fi

\begin{document}

\maketitle

\begin{abstract}
Search is a central component of \gls*{ASR} systems and has been deeply investigated for classical ASR modeling based on \glspl*{HMM}. However, in work on current sequence-to-sequence ASR architectures it is often simplified to a small fixed-size beam search without external \gls*{LM}.
In this work, we present a comparative analysis of search space complexity and efficiency for \gls*{CTC} and neural transducer models under different pruning strategies and \gls*{LM} configurations.
We show that these \gls*{E2E} models work with a small search space most of the time but exhibit large temporary peaks in hypothesis count during periods of uncertainty.
Score-based pruning dynamically adapts the search to model uncertainty, leading to a significantly lower \gls*{RTF}.
Our results highlight the limitations of search with low fixed-size beam pruning and provide practical guidelines for efficient search in \gls*{E2E} \gls*{ASR}.

\end{abstract}


\section{Introduction \& Related Work}
\label{sec:intro}
Search (or decoding) is the central process that brings together the components of an \gls*{ASR} system in order to turn acoustic evidence into a transcription by navigating a space of intermediate hypotheses over time.
In the \gls*{LVCSR} era of the 1990s-2000s, search spaces often contained thousands or tens of thousands of hypotheses per time step, which drove the need for efficient search strategies and methods for search space reduction \cite{steinbiss1994improvements, aubert2002overview, wendt2002dynamic, pylkkonen2005new}.
Modern neural \gls*{ASR} models such as \gls*{CTC} \cite{graves2006ctc}, neural transducers \cite{graves2012rnnt} and \gls*{AED} \cite{chorovski2015aed} models have changed this setting.
They often exhibit sharp posterior distributions where most of the time the probability mass is concentrated on only one output class \cite{zeyer2021peakyctc}.
Additionally, these models are commonly defined on an orthographic (e.g. character or \gls*{BPE} \cite{gage1994bpe}) rather than a phonetic subword vocabulary, making search possible without a pronunciation lexicon or external \gls*{LM}.
Due to this development, more sophisticated search methods have been largely disregarded in recent literature, and many works just report fixed beam-sizes of 1-8 without external \gls*{LM} integration.
Dynamic score-threshold pruning and lexical prefix-tree based search organization are known techniques from classical \gls*{ASR}, but their practical role for modern time-synchronous \gls*{E2E} architectures has been underrepresented.

Most publicly available toolkits reflect this shift of focus away from search and only offer limited search capabilities for modern model architectures.
Flashlight \cite{kahn2022flashlight} offers lexicon-constrained and lexicon-free search as well as word- and subword-level \gls*{LM} integration, but separate pruning of within-word and word-end hypotheses is not possible and it does not provide support for transducer models.
ESPnet \cite{watanabe2018espnet} provides transducer search with neural \glspl*{LM} at the subword-level but does not provide lexicon- or word-level capabilities and only fixed-size beam pruning instead of dynamic score-based pruning.
k2/icefall \cite{povey2011kaldi} also offers many options for \gls*{CTC} search, including a lexicon and n-gram \gls*{LM} but only implements neural \glspl*{LM} via rescoring and again lacks dynamic pruning based on a score threshold for transducer search.
NeMo \cite{kuchaiev2019nemo} and recent publications with it \cite{xu2025wind, bataev2025ngpu, grigoryan2025rnntbeam} aim to optimize the transducer search process fully on GPU in order to take advantage of parallelization and avoid inter-device communication.
However, this approach again prohibits word-level processing and dynamic pruning.

In this paper, we challenge the assumption that small fixed-size beam search is sufficient for optimal performance with modern model architectures, in terms of both accuracy and speed.
In our open-source toolkit \toolkitname{} \ifanonymous \else \cite{wiesler2014rasr, rybach2011rasr, zhou2023rasr2} \fi \footnote{\ifanonymous Anonymized for double-blind review \else \url{https://github.com/rwth-i6/rasr} \fi} we implemented a generic search module which supports various search strategies for various \gls*{E2E} \gls*{ASR} models, allowing a comprehensive study of their behavior and search space under comparable conditions.
We restrict the analysis in this work to time-synchronous \gls*{CTC} and monotonic transducer models because \gls*{AED}, Speech-\gls*{LLM} and related sequence-to-sequence models search label-synchronously which means that hypotheses at one search step may correspond to different input positions and finished hypotheses may have different lengths.
This makes active-hypothesis-count and pruning statistics not directly comparable and introduces the length-normalization problem which is out of scope for this discussion.

The main contributions of our work are as follows:
\begin{itemize}
	\item We systematically compare the performance of \gls*{BPE} and phoneme-based \gls*{CTC} and neural transducer models under various search conditions such as with and without lexicon-constraint as well as with and without subword- or word-level \glspl*{LM}.
	\item We analyze search space behavior across different models and task difficulty levels by computing search errors and hypothesis counts.
    \item We quantify the impact of dynamic vs. fixed-size beam pruning on the search accuracy (\gls*{WER}) and efficiency (\gls*{RTF}) across datasets and search conditions.
    \item We give compact algorithmic descriptions and tuned decoding parameters for reproducibility.
\end{itemize}
Our setups are published on GitHub\footnote{\ifanonymous Anonymized for double-blind review \else \url{https://github.com/rwth-i6/returnn-experiments/tree/master/2026-search-space} \fi}.

\section{Background}
\label{sec:background}
\begin{algorithm}[t]
    \caption{One step of lexicon-free time-synchronous beam search. Scores are
    negative log probabilities. The subword-level LM scorer and the corresponding
    intermediate pruning step are optional.}
    \label{alg:lexfree-search}
    \begin{algorithmic}[1]
        \Require beam $\mathcal{H}$, output labels $\mathcal{A}$, ASR score
        function $\mathrm{score}_{\mathrm{ASR}}$, optional subword-LM score
        function $\mathrm{score}_{\mathrm{SubLM}}$
        \Require score thresholds $\Delta_{\mathrm{ASR}},\Delta_{\mathrm{SubLM}}$,
        intermediate max beam size $B_{\mathrm{ASR}}$, final max beam size $B_{\mathrm{SubLM}}$
        \State $\mathcal{E} \gets \emptyset$
        \ForAll{$q\in\mathcal{H}$ and $a\in\mathcal{A}$}
            \State $\mathrm{score}\gets
            \mathrm{score}(q)+\mathrm{score}_{\mathrm{ASR}}(q,a)$
            \State $\mathcal{E}\gets\mathcal{E}\cup\{(q,a,\mathrm{score})\}$
        \EndFor
        \State $\mathcal{E}\gets
        \Call{ScorePrune}{\mathcal{E},\Delta_{\mathrm{ASR}}}$
        \If{$\mathrm{LM}$ is present}
            \State $\mathcal{E}\gets
            \Call{BeamPrune}{\mathcal{E},B_{\mathrm{ASR}}}$
            \ForAll{$e=(q,a,\mathrm{score})\in\mathcal{E}$}
                \If{$a\neq\langle\mathrm{blank}\rangle$}
                    \State $\mathrm{score}(e)\gets\mathrm{score}(e)
                    +\lambda \cdot \mathrm{score}_{\mathrm{SubLM}}(q,a)$
                \EndIf
            \EndFor
            \State $\mathcal{E}\gets
            \Call{ScorePrune}{\mathcal{E},\Delta_{\mathrm{SubLM}}}$
        \EndIf
        \State $\mathcal{H}'\gets
        \{\Call{Extend}{q,a,\mathrm{score}}:(q,a,\mathrm{score})\in\mathcal{E}\}$
        \State $\mathcal{H}'\gets\Call{Recombine}{\mathcal{H}'}$
        \State $\mathcal{H}'\gets\Call{BeamPrune}{\mathcal{H}',B_{\mathrm{SubLM}}}$
        \State \Return $\mathcal{H}'$
    \end{algorithmic}
\end{algorithm}

\begin{algorithm}[t]
    \caption{One step of lexicon-constrained time-synchronous tree search.
    Scores are negative log probabilities. Subword- and word-LM scoring
    are optional; word ends update the word-LM history and return to a
    tree root.}
    \label{alg:tree-search}
    \begin{algorithmic}[1]
        \Require beam $\mathcal{H}$, lexical prefix tree $\mathcal{T}$,
        subword-level ASR score $\mathrm{score}_{\mathrm{ASR}}$, optional
        subword-LM score $\mathrm{score}_{\mathrm{SubLM}}$, optional
        word-LM score $\mathrm{score}_{\mathrm{WordLM}}$
        \Require score thresholds $\Delta_{\mathrm{ASR}},
        \Delta_{\mathrm{SubLM}},\Delta_{\mathrm{WordLM}}$,
        intermediate max beam size $B_{\mathrm{ASR}}$, subword-level max beam size
        $B_{\mathrm{SubLM}}$, word-level max beam size $B_{\mathrm{WordLM}}$
        \State $\mathcal{E}_{\mathrm{Subword}}\gets\emptyset$
        \ForAll{$q\in\mathcal{H}$ and $(a,s_{\mathrm{next}})\in
        \Call{Successors}{\mathcal{T},\mathrm{state}(q)}$}
            \State $\mathrm{score}\gets
            \mathrm{score}(q)+\mathrm{score}_{\mathrm{ASR}}(q,a)$
            \State $\mathcal{E}_{\mathrm{Subword}}\gets
            \mathcal{E}_{\mathrm{Subword}}\cup
            \{(q,a,s_{\mathrm{next}},\mathrm{score})\}$
        \EndFor
        \State $\mathcal{E}_{\mathrm{Subword}}\gets
        \Call{ScorePrune}{\mathcal{E}_{\mathrm{Subword}},\Delta_{\mathrm{ASR}}}$
        \If{$\mathrm{SubLM}$ is present}
            \State $\mathcal{E}_{\mathrm{Subword}}\gets
            \Call{BeamPrune}{\mathcal{E}_{\mathrm{Subword}},B_{\mathrm{ASR}}}$
            \ForAll{$e=(q,a,s_{\mathrm{next}},\mathrm{score})
            \in\mathcal{E}_{\mathrm{Subword}}$}
                \If{$a\neq\langle\mathrm{blank}\rangle$}
                    \State $\mathrm{score}(e)\gets\mathrm{score}(e)
                    +\lambda_2\cdot\mathrm{score}_{\mathrm{SubLM}}(q,a)$
                \EndIf
            \EndFor
            \State $\mathcal{E}_{\mathrm{Subword}}\gets
            \Call{ScorePrune}{\mathcal{E}_{\mathrm{Subword}},
            \Delta_{\mathrm{SubLM}}}$
        \EndIf
        \State $\mathcal{H}_{\mathrm{Subword}}\gets
        \{\Call{ExtendSubword}{e}:e\in\mathcal{E}_{\mathrm{Subword}}\}$
        \State $\mathcal{H}_{\mathrm{Subword}}\gets
        \Call{RecombineSubword}{\mathcal{H}_{\mathrm{Subword}}}$
        \State $\mathcal{H}_{\mathrm{Subword}}\gets
        \Call{BeamPrune}{\mathcal{H}_{\mathrm{Subword}},B_{\mathrm{SubLM}}}$
        \State $\mathcal{E}_{\mathrm{word}}\gets\emptyset$
        \ForAll{$q\in\mathcal{H}_{\mathrm{Subword}}$ and
        $(w,s_{\mathrm{root}})\in
        \Call{WordEnds}{\mathcal{T},\mathrm{state}(q)}$}
            \State $\mathrm{score}\gets \mathrm{score}(q)$
            \If{$\mathrm{WordLM}$ is present}
                \State $\mathrm{score}\mathrel{+}=
                \lambda_1\cdot\mathrm{score}_{\mathrm{WordLM}}(q,w)$
            \EndIf
            \State $\mathcal{E}_{\mathrm{word}}\gets
            \mathcal{E}_{\mathrm{word}}\cup
            \{(q,w,s_{\mathrm{root}},\mathrm{score})\}$
        \EndFor
        \State $\mathcal{E}_{\mathrm{word}}\gets
        \Call{ScorePrune}{\mathcal{E}_{\mathrm{word}},
        \Delta_{\mathrm{WordLM}}}$
        \State $\mathcal{H}_{\mathrm{word}}\gets
        \{\Call{ExtendWordEnd}{e}:e\in\mathcal{E}_{\mathrm{word}}\}$
        \State $\mathcal{H}_{\mathrm{word}}\gets
        \Call{RecombineWordEnd}{\mathcal{H}_{\mathrm{word}}}$
        \State $\mathcal{H}_{\mathrm{word}}\gets
        \Call{BeamPrune}{\mathcal{H}_{\mathrm{word}},B_{\mathrm{WordLM}}}$
        \State \Return $\mathcal{H}_{\mathrm{Subword}}
        \cup\mathcal{H}_{\mathrm{word}}$
    \end{algorithmic}
\end{algorithm}

In modern \gls*{ASR}, the search maps a sequence of encoder states $h_1^T$ which are computed from acoustic features $x_1^{T'}$ to the output sequence with maximum probability.
This can be conducted in a lexicon-free (open-vocabulary) or a lexicon-constrained (closed-vocabulary) manner.
In the lexicon-free case, the maximization is carried out directly over subword sequences $a_1^S$:
\begin{equation} \label{eq:lexfreebayes}
    h_1^T \rightarrow \argmax_{a_1^S} \bigg\{ p_\mathrm{SubLM}(a_1^S)^\lambda \cdot \max_{y_1^T : a_1^S} p_\mathrm{ASR}(y_1^T \mid h_1^T)\bigg\}.
\end{equation}
Here, $y_1^T : a_1^S$ denotes the set of alignments allowed by the label topology that map or collapse to the subword sequence $a_1^S$.
For each subword sequence there is a score contribution from a subword \gls*{LM} $p_\mathrm{SubLM}$ scaled by a factor $\lambda$ (which may be 0) and an \gls*{ASR} model $p_\mathrm{ASR}$ which is broken down to frame-wise alignments.

Lexicon-constrained search works in a similar way but has an additional outer maximization over word sequences $w_1^N$:
\begin{equation} \label{eq:treebayes}
		\begin{split}
				h_1^T \rightarrow \argmax_{w_1^N} &\bigg\{ p_\mathrm{WordLM}(w_1^N)^{\lambda_1} \cdot \max_{a_1^S : w_1^N} \Big\{ p_\mathrm{SubLM}(a_1^S)^{\lambda_2} \\
				&\cdot \max_{y_1^T : a_1^S} p_\mathrm{ASR}(y_1^T \mid h_1^T) \Big\} \bigg\}.
		\end{split}
\end{equation}
In the inner maximization, only those label sequences $a_1^S$ that correspond to a valid word sequence according to the lexicon are considered.
This type of search also allows the integration of a word-level \gls*{LM} $p_\mathrm{WordLM}$ in addition to $p_\mathrm{SubLM}$.
In practice, however, usually only a word-level \gls*{LM} is used during lexicon-constrained search.

In this work, we focus on \gls*{CTC} and monotonic neural transducer \cite{tripathi2019monotonic} \gls*{ASR} models, since both of them work on a frame-by-frame basis and allow for a meaningful comparison.
In the case of \gls*{CTC}, the alignment probability in search can be stated as follows, including a frame-wise prior probability $p_\mathrm{Prior}(y_t)$ with scale $\beta$:
\begin{equation} \label{eq:ctc}
    p_\mathrm{ASR}(y_1^T \mid h_1^T) = \prod_{t=1}^T \frac{p_\mathrm{CTC}(y_t \mid h_t)}{p_\mathrm{Prior}(y_t)^\beta}.
\end{equation}
For the transducer model, we use a limited context of one history label, which has previously shown to yield performance equivalent to full context \cite{zhou2021phonemetransducer}.
We also correct by the \gls*{ILM} probability $p_\mathrm{ILM}(a_s \mid a_{s-1})$ which we estimate by the zero-encoder method \cite{variani2020hat}.
For convenience of notation, we define $a(y_1^{t-1})$ as the last non-blank label in $y_1^{t-1}$ and set $p_\mathrm{ILM}(y \mid a) = 1$ when $y$ is the blank label since the \gls*{ILM} only contributes for non-blank labels:
\begin{equation} \label{eq:transducer}
    p_\mathrm{ASR}(y_1^T \mid h_1^T) = \prod_{t=1}^T \frac{p_\mathrm{Transducer}(y_t \mid a(y_1^{t-1}), h_t)}{p_\mathrm{ILM}(y_t \mid a(y_1^{t-1}))^\beta}.
\end{equation}

\Cref{alg:lexfree-search,alg:tree-search} implement the global objectives defined in equations \eqref{eq:lexfreebayes} and \eqref{eq:treebayes} approximately in negative-log space by accumulating local \gls*{ASR} and \gls*{LM} scores.
Here, $\mathrm{score}_{\mathrm{ASR}} = - \log p_{\mathrm{ASR}}$ is instantiated by \Cref{eq:ctc} for \gls*{CTC} and \Cref{eq:transducer} for the transducer.
Recombination keeps the best hypothesis among equivalent contexts.
For lexicon-free search, this recombination context consists of prediction network state (for transducer) and the subword-\gls*{LM} history.
For lexicon-constrained search, the recombination context additionally contains tree state and the word-\gls*{LM} history.

During the search process, we apply score-based and max-beam-size-based pruning strategies.
In score-based pruning with a threshold $\Delta$, the current best hypothesis is selected from the beam and all hypotheses whose score (i.e., negative log probability) is worse than the current best plus the threshold are pruned.
Let $\mathcal{H}$ be the current set of hypotheses, then this is formalized as:
\begin{equation} \label{eq:scorepruning}
        \textsc{ScorePrune}(\mathcal{H}, \Delta)
        = \Big\{ q \in \mathcal{H} \mid {}
        \mathrm{score}(q) \leq
        \min_{q' \in \mathcal{H}}\lbrace \mathrm{score}(q')\rbrace
        + \Delta \Big\}
\end{equation}
On the other hand, with max-beam-size pruning, we simply keep a fixed maximum amount $B$ of hypotheses while pruning away the rest, i.e.:
\begin{equation} \label{eq:beamsizepruning}
    \textsc{BeamPrune}(\mathcal{H}, B) = \mathrm{Top}_B(\mathcal{H})
\end{equation}

Pure fixed-beam search disables \textsc{ScorePrune} and only applies \textsc{BeamPrune}.
In our search implementation, we apply score- and max-beam-size-pruning with the option to use different thresholds after each individual score contribution, i.e., first after $\mathrm{score}_\mathrm{ASR}$, then after $\mathrm{score}_\mathrm{SubLM}$ and finally after $\mathrm{score}_\mathrm{WordLM}$.
We also prune within-word hypotheses separately from word-end hypotheses.
We denote the ASR-stage, subword-\gls*{LM} and word-\gls*{LM} maximum beam sizes by $B_\mathrm{ASR}$, $B_\mathrm{SubLM}$ and $B_\mathrm{WordLM}$, and their score thresholds by $\Delta_\mathrm{ASR}$, $\Delta_\mathrm{SubLM}$ and $\Delta_\mathrm{WordLM}$, respectively.

All decoding experiments use \toolkitname{} as the implementation framework.
It supports, among others, \gls*{CTC} and transducer scorers, subword- and word-level \glspl*{LM}, lexicon-free and lexicon-constrained decoding, and the staged pruning described above.
\Cref{alg:lexfree-search,alg:tree-search} summarize the corresponding search steps implementations, but omit some optimizations such as on-the-fly extension pruning.
Notably, all results characterize the model and search procedure, not \toolkitname{}-specific behavior.

\section{Experimental Setup}
\label{sec:setup}
Experiments are conducted on three corpora: LibriSpeech, Loquacious, and an in-house Spanish telephony corpus.
LibriSpeech \cite{panayotov2015librispeech} uses the full 960h training set and is evaluated on dev-other and test-other.
For Loquacious, we use the public medium portion with 2500h of 16kHz training data \cite{parcollet25loquacious}.
Loquacious evaluation is performed on dev.all and test.all, which aggregate VoxPopuli, LibriSpeech, YODAS, and Common Voice.
The Spanish corpus contains around 12000h of 8kHz telephony training data and is evaluated on an in-house 4.5h test set, providing a harder-domain cross-lingual condition.

We train models on \gls*{BPE} and phoneme alphabets.
The LibriSpeech \gls*{BPE} vocabulary contains 185 labels including blank while the \gls*{EOW} augmented phoneme vocabulary contains 80 labels including blank.
The Loquacious setup uses \gls*{BPE}/\gls*{EOW}-phoneme vocabularies of 184/80 labels including blank.
On the Spanish task, we use a vocabulary of 63 \gls*{EOW}-phoneme labels.

For acoustic features, we extract 80-dimensional log-mel features from the 16kHz LibriSpeech and Loquacious data, and 60-dimensional log-mel features from the 8kHz Spanish data.
The features are passed through a convolutional front-end and subsampled to a step-size of 40 ms.
Afterwards, they are processed by a Conformer \cite{gulati2020conformer} encoder with 12 layers for LibriSpeech and Loquacious, and 16 layers for Spanish.
The transducer models additionally use a 2-layer feed-forward prediction network for the limited history and a 1-layer feed-forward joiner network.

Each LibriSpeech model is trained for 100 epochs, each Loquacious model for 40 epochs, and the Spanish model for 60 epochs.
All models use a \gls*{OCLR} \cite{smith2019oclr} schedule and apply SpecAugment \cite{park2019specaugment}, speed perturbation \cite{ko2015speedperturbation}, dropout, and weight decay for regularization.

For word-level \gls*{LM} experiments, we apply a 4-gram \gls*{LM}; for LibriSpeech, this is the official 4-gram \gls*{LM}, for Loquacious, this is the 4-gram \gls*{LM} setup described in \cite{rossenbach2026loquacious}, and for Spanish, it is trained on in-house data.
On LibriSpeech, we additionally present results with a small 2-layer \gls*{LSTM} \cite{hochreiter1997lstm} \gls*{LM} with 55M parameters at the \gls*{BPE} level and larger 96-layer Transformer \glspl*{LM} at \gls*{BPE} and word level with around 300M and 430M parameters respectively \cite{irie2019trafolm}.

\section{Search Strategies}
\label{sec:searchstrategies}
\begin{table}
\centering
\caption{WER and RTF results on LibriSpeech for different models given various search conditions and LM combinations.}
\label{tab:wers_librispeech}
\begin{tabular}{|c|c|c|c|r|r|r|}
\hline
\multirow{2}{*}{Model}                                                & \multirow{2}{*}{Lex.} & \multicolumn{2}{c|}{LM} & \multicolumn{2}{c|}{WER [\%]}                            & \multicolumn{1}{c|}{\multirow{2}{*}{RTF $\cdot 10^3$}}  \\ 
\cline{3-6}
                                                                      &                       & Subword & Word          & \multicolumn{1}{c|}{dev-o} & \multicolumn{1}{c|}{test-o} & \multicolumn{1}{c|}{}                                                                      \\ 
\hhline{|=======|}
\multirow{8}{*}{\begin{tabular}[c]{@{}c@{}}BPE\\CTC\end{tabular}}     & \multirow{3}{*}{No}   & -       & -             & 7.0                        & 7.1                         & 0.62                                                                                      \\ 
\cline{3-7}
                                                                      &                       & LSTM    & -             & 5.1                        & 5.1                         & 22                                                                                      \\ 
\cline{3-7}
                                                                      &                       & TraFo   &               & 4.6                        & 4.7                         & 3357                                                                                      \\ 
\cline{2-7}
                                                                      & \multirow{5}{*}{Yes}  & -       & -             & 6.5                        & 6.6                         & 0.93                                                                                      \\ 
\cline{3-7}
                                                                      &                       & -       & 4-gram         & 5.2                        & 5.6                         & 1.6                                                                                      \\ 
\cline{3-7}
                                                                      &                       & LSTM    & -             & 5.2                        & 5.3                         & 9.0                                                                                      \\ 
\cline{3-7}
                                                                      &                       & LSTM    & 4-gram         & 5.1                        & 5.3                         & 7.5                                                                                      \\ 
\cline{3-7}
                                                                      &                       & -       & TraFo         & 4.6                        & 4.7                         & 154                                                                                      \\ 
\hhline{|=======|}
\multirow{2}{*}{\begin{tabular}[c]{@{}c@{}}Phon.\\CTC\end{tabular}}   & \multirow{2}{*}{Yes}  & -       & 4-gram         & 5.5                        & 5.7                         & 1.5                                                                                      \\ 
\cline{3-7}
                                                                      &                       & -       & TraFo         & 4.5                        & 4.8                         & 290                                                                                      \\ 
\hhline{|=======|}
\multirow{8}{*}{\begin{tabular}[c]{@{}c@{}}BPE\\Transd.\end{tabular}} & \multirow{3}{*}{No}   & -       & -             & 6.0                        & 5.9                         & 7.6                                                                                      \\ 
\cline{3-7}
                                                                      &                       & LSTM    & -             & 4.5                        & 4.8                         & 27                                                                                      \\ 
\cline{3-7}
                                                                      &                       & TraFo   & -             & 4.1                        & 4.3                         & 3488                                                                     \\ 
\cline{2-7}
                                                                      & \multirow{5}{*}{Yes}  & -       & -             & 5.9                        & 5.9                         & 8.1                                                                                      \\ 
\cline{3-7}
                                                                      &                       & -       & 4-gram         & 4.8                        & 5.1                         & 13                                                                                      \\ 
\cline{3-7}
                                                                      &                       & LSTM    & -             & 4.7                        & 5.1                         & 21                                                                                      \\ 
\cline{3-7}
                                                                      &                       & LSTM    & 4-gram         & 4.8                        & 5.1                         & 20                                                                                      \\
\cline{3-7}
                                                                      &                       & -       & TraFo         & 3.8                        & 4.2                         & 310                                                                                      \\
\hline
\end{tabular}
\end{table}

\begin{table}
\centering
\caption{WER and RTF results on Loquacious for different models given various search conditions and LM combinations.}
\label{tab:wers_loquacious}
\begin{tabular}{|c|c|c|r|r|r|}
\hline
\multirow{2}{*}{Model}                                                & \multirow{2}{*}{Lex.}                     & \multirow{2}{*}{Word LM} & \multicolumn{2}{c|}{WER [\%]}                                & \multicolumn{1}{c|}{\multirow{2}{*}{RTF $\cdot 10^3$}}  \\ 
\cline{4-5}
                                                                      &                                           &                          & \multicolumn{1}{c|}{dev.all} & \multicolumn{1}{c|}{test.all} & \multicolumn{1}{c|}{}                                                                      \\ 
\hhline{|======|}
\multirow{3}{*}{\begin{tabular}[c]{@{}c@{}}BPE\\CTC\end{tabular}}     & No                                        & -                        & 11.7                         & 12.8                          & 0.62                                                                                      \\ 
\cline{2-6}
                                                                      & \multirow{2}{*}{Yes}                      & -                        & 11.0                         & 12.0                          & 0.73                                                                                      \\ 
\cline{3-6}
                                                                      &                                           & 4-gram                    & 9.5                          & 10.4                          & 0.89                                                                                      \\ 
\hhline{|======|}
\multirow{3}{*}{\begin{tabular}[c]{@{}c@{}}BPE\\Transd.\end{tabular}} & No                                        & -                        & 9.0                          & 10.2                          & 7.7                                                                                      \\ 
\cline{2-6}
                                                                      & \multicolumn{1}{l|}{\multirow{2}{*}{Yes}} & -                        & 9.0                          & 10.1                          & 8.6                                                                                      \\ 
\cline{3-6}
                                                                      & \multicolumn{1}{l|}{}                     & 4-gram                    & 8.0                          & 8.9                           & 13                                                                                      \\ 
\hhline{|======|}
\begin{tabular}[c]{@{}c@{}}Phon.\\Transd.\end{tabular}                & Yes                                       & 4-gram                    & 8.4                          & 9.2                           & 13                                                                                      \\
\hline
\end{tabular}
\end{table}

\Cref{tab:wers_librispeech} shows \gls*{WER} and search \gls*{RTF} results for the LibriSpeech models under different search conditions and \gls*{LM} integrations.
The search \gls*{RTF} only measures the computational speed of the search process, which includes hypothesis extension, predictor and joiner forwarding for transducer, \gls*{LM} forwarding, pruning and bookkeeping but excludes the encoder forwarding.
It is measured with all neural components (i.e., neural \glspl*{LM} and transducer predictor and joiner) running on an A10 GPU while the search logic and 4-gram \gls*{LM} run on CPU (note: the encoder \gls*{RTF} on an A10 is about $3.0 \times 10^{-3}$).
The presented results cover decoding with no external \gls*{LM}, subword-level neural \gls*{LSTM} or Transformer \gls*{LM} integration, and word-level Transformer-\gls*{LM} or 4-gram integration.
Lexicon constraints improve \gls*{CTC} without an external \gls*{LM} (from 7.0/7.1\% to 6.5/6.6\% \gls*{WER}) and have little effect for the transducer.
All external-\gls*{LM} combinations outperform the standalone \gls*{ASR} models by a good margin, with the best setup reaching 3.8/4.2\% \gls*{WER}.
Overall, the word-level Transformer \gls*{LM} gives the best \gls*{WER}, while the 4-gram and subword-level \gls*{LSTM} \glspl*{LM} provide decent accuracy at much higher speed.
The subword-level Transformer \gls*{LM} is expensive in terms of execution time because it needs to score every (non-blank) subword extension, while not improving \gls*{WER} over the word-level Transformer \gls*{LM} which only needs to be forwarded at word ends.
The subword-level \gls*{LSTM} \gls*{LM} slightly improves over the 4-gram \gls*{LM}, but is also slower; combining the \gls*{LSTM} and 4-gram \glspl*{LM} gives results similar to the \gls*{LSTM} alone.
A natural direction for future work is to extend this comparison to larger \gls*{ASR} models trained on more data, where integration of external \glspl*{LM} may have lower impact.

The phoneme \gls*{CTC} model is slightly worse than \gls*{BPE} \gls*{CTC} with a 4-gram \gls*{LM} (5.5/5.7\% vs. 5.2/5.6\%) and on par with the Transformer \gls*{LM}.
Search without a lexicon and word-\gls*{LM} is not applicable to phoneme models since phonetically equivalent words cannot be distinguished.
The transducer outperforms the \gls*{CTC} models, especially when no lexicon or \gls*{LM} is present (6.0/5.9\% vs. 7.0/7.1\% \gls*{WER}) and remains better by between 0.3\% to 0.7\% absolute \gls*{WER} in all other conditions.
Compared to \gls*{CTC}, transducer search is much slower without \gls*{LM} or with only a 4-gram \gls*{LM}, but the gap shrinks with a neural \gls*{LM} because \gls*{LM} forwarding makes up a large part of the search cost.

\Cref{tab:wers_loquacious} reports the same core search conditions without neural \glspl*{LM} on Loquacious, providing evidence on a public heterogeneous dataset beyond LibriSpeech.

\section{Search Space Complexity}
\label{sec:searchspace}
\begin{table}
\centering
\caption{WER and hypothesis count statistics for different pruning parameters with BPE CTC lexicon-constrained search + word-level Transformer LM on LibriSpeech dev-other.}
\label{tab:bpe_ctc_search_space}
\begin{tabular}{|r|r|r|r|r|r|r|} 
\hline
\multicolumn{1}{|c|}{\multirow{2}{*}{\begin{tabular}[c]{@{}c@{}}Score\\Thresh.\end{tabular}}} & \multicolumn{1}{c|}{\multirow{2}{*}{\begin{tabular}[c]{@{}c@{}}Beam\\Size\end{tabular}}} & \multicolumn{1}{c|}{\multirow{2}{*}{\begin{tabular}[c]{@{}c@{}}WER\\{[}\%]\end{tabular}}} & \multicolumn{1}{c|}{\multirow{2}{*}{\begin{tabular}[c]{@{}c@{}}Search\\Err. [\%]\end{tabular}}} & \multicolumn{3}{c|}{\#Hypotheses}                                               \\ 
\cline{5-7}
\multicolumn{1}{|c|}{}                                                                        & \multicolumn{1}{c|}{}                                                                    & \multicolumn{1}{c|}{}                                                                     & \multicolumn{1}{c|}{}                                                                           & \multicolumn{1}{c|}{p50} & \multicolumn{1}{c|}{p90} & \multicolumn{1}{c|}{p99}  \\ 
\hhline{|=======|}
0.0                                                                                           & 1                                                                                        & 10.4                                                                                      & 30.4                                                                                            & 1                        & 1                        & 1                         \\ 
\hline
4.0                                                                                           & 16                                                                                       & 9.4                                                                                       & 14.7                                                                                            & 1                        & 2                        & 7                         \\ 
\hline
8.0                                                                                           & 128                                                                                      & 6.7                                                                                       & 4.9                                                                                             & 2                        & 6                        & 41                        \\ 
\hline
12.0                                                                                          & 256                                                                                      & 4.8                                                                                       & 1.3                                                                                             & 5                        & 28                       & 242                       \\ 
\hline
16.0                                                                                          & 256                                                                                      & 4.6                                                                                       & 0.6                                                                                             & 35                       & 167                      & 254                       \\
\hline
\end{tabular}
\end{table}

\begin{table}
\centering
\caption{WER and hypothesis count statistics for different pruning parameters with BPE transducer lexicon-constrained search + word-level Transformer LM on LibriSpeech dev-other.}
\label{tab:bpe_transducer_search_space}
\begin{tabular}{|r|r|r|r|r|r|r|} 
\hline
\multicolumn{1}{|c|}{\multirow{2}{*}{\begin{tabular}[c]{@{}c@{}}Score\\Thresh.\end{tabular}}} & \multicolumn{1}{c|}{\multirow{2}{*}{\begin{tabular}[c]{@{}c@{}}Beam\\Size\end{tabular}}} & \multicolumn{1}{c|}{\multirow{2}{*}{\begin{tabular}[c]{@{}c@{}}WER\\{[}\%]\end{tabular}}} & \multicolumn{1}{c|}{\multirow{2}{*}{\begin{tabular}[c]{@{}c@{}}Search\\Err. [\%]\end{tabular}}} & \multicolumn{3}{c|}{\#Hypotheses}                                               \\ 
\cline{5-7}
\multicolumn{1}{|c|}{}                                                                        & \multicolumn{1}{c|}{}                                                                    & \multicolumn{1}{c|}{}                                                                     & \multicolumn{1}{c|}{}                                                                           & \multicolumn{1}{c|}{p50} & \multicolumn{1}{c|}{p90} & \multicolumn{1}{c|}{p99}  \\ 
\hhline{|=======|}
0.0                                                                                           & 1                                                                                        & 10.0                                                                                      & 24.1                                                                                            & 1                        & 1                        & 1                         \\ 
\hline
4.0                                                                                           & 16                                                                                       & 11.0                                                                                      & 20.3                                                                                            & 1                        & 2                        & 6                         \\ 
\hline
8.0                                                                                           & 128                                                                                      & 6.9                                                                                       & 7.2                                                                                             & 2                        & 7                        & 51                        \\ 
\hline
12.0                                                                                          & 256                                                                                      & 4.2                                                                                       & 1.7                                                                                             & 7                        & 46                       & 250                       \\ 
\hline
16.0                                                                                          & 1024                                                                                     & 3.8                                                                                       & 0.4                                                                                             & 55                       & 433                      & 1012                      \\
\hline
\end{tabular}
\end{table}

\begin{table}
\centering
\caption{WER and hypothesis count statistics for different pruning parameters with BPE transducer lexicon-constrained search + word-level 4-gram LM on Loquacious dev.all.}
\label{tab:bpe_transducer_search_space_loquacious}
\begin{tabular}{|r|r|r|r|r|r|r|} 
\hline
\multicolumn{1}{|c|}{\multirow{2}{*}{\begin{tabular}[c]{@{}c@{}}Score\\Thresh.\end{tabular}}} & \multicolumn{1}{c|}{\multirow{2}{*}{\begin{tabular}[c]{@{}c@{}}Beam\\Size\end{tabular}}} & \multicolumn{1}{c|}{\multirow{2}{*}{\begin{tabular}[c]{@{}c@{}}WER\\{[}\%]\end{tabular}}} & \multicolumn{1}{c|}{\multirow{2}{*}{\begin{tabular}[c]{@{}c@{}}Search\\Err. [\%]\end{tabular}}} & \multicolumn{3}{c|}{\#Hypotheses}                                               \\ 
\cline{5-7}
\multicolumn{1}{|c|}{}                                                                        & \multicolumn{1}{c|}{}                                                                    & \multicolumn{1}{c|}{}                                                                     & \multicolumn{1}{c|}{}                                                                           & \multicolumn{1}{c|}{p50} & \multicolumn{1}{c|}{p90} & \multicolumn{1}{c|}{p99}  \\ 
\hhline{|=======|}
0.0                                                                                           & 1                                                                                        & 11.9                                                                                      & 16.8                                                                                            & 1                        & 1                        & 1                         \\ 
\hline
2.0                                                                                           & 16                                                                                       & 10.9                                                                                      & 10.9                                                                                            & 1                        & 2                        & 4                         \\ 
\hline
4.0                                                                                           & 32                                                                                       & 8.6                                                                                       & 2.6                                                                                             & 1                        & 3                        & 11                        \\ 
\hline
6.0                                                                                           & 32                                                                                       & 8.0                                                                                       & 0.7                                                                                             & 1                        & 5                        & 30                        \\ 
\hline
8.0                                                                                           & 32                                                                                       & 7.9                                                                                       & 0.5                                                                                             & 2                        & 11                       & 31                        \\
\hline
\end{tabular}
\end{table}

\begin{table}
\centering
\caption{WER and hypothesis count statistics for different pruning parameters with phoneme CTC lexicon-constrained search + word-level 4-gram LM on Spanish telephony.}
\label{tab:phoneme_ctc_search_space}
\begin{tabular}{|r|r|r|r|r|r|r|} 
\hline
\multicolumn{1}{|c|}{\multirow{2}{*}{\begin{tabular}[c]{@{}c@{}}Score\\Thresh.\end{tabular}}} & \multicolumn{1}{c|}{\multirow{2}{*}{\begin{tabular}[c]{@{}c@{}}Beam\\Size\end{tabular}}} & \multicolumn{1}{c|}{\multirow{2}{*}{\begin{tabular}[c]{@{}c@{}}WER\\{[}\%]\end{tabular}}} & \multicolumn{4}{c|}{\#Hypotheses}                                                                           \\ 
\cline{4-7}
\multicolumn{1}{|c|}{}                                                                        & \multicolumn{1}{c|}{}                                                                    & \multicolumn{1}{c|}{}                                                                     & \multicolumn{1}{c|}{Avg.} & \multicolumn{1}{c|}{p50} & \multicolumn{1}{c|}{p90} & \multicolumn{1}{c|}{p99}  \\ 
\hhline{|=======|}
0.0                                                                                           & 1                                                                                        & 46.6                                                                                      & 1.0                       & 1                        & 1                        & 1                         \\ 
\hline
4.0                                                                                           & 128                                                                                      & 17.9                                                                                      & 6.1                       & 3                        & 11                       & 79                        \\ 
\hline
6.0                                                                                           & 512                                                                                      & 15.5                                                                                      & 88.3                      & 55                       & 180                      & 512                       \\ 
\hline
8.0                                                                                           & 1024                                                                                     & 15.0                                                                                      & 251.0                     & 148                      & 675                      & 1024                      \\ 
\hline
10.0                                                                                          & 4096                                                                                     & 14.9                                                                                      & 932.2                     & 451                      & 2926                     & 4096                      \\
\hline
\end{tabular}
\end{table}

We first investigate the relation between pruning parameters, search space size, and recognition performance.
To optimize the pruning parameters, we first set a very large maximum beam-size $B$ while tuning the score thresholds and then decrease $B$ until performance starts to degrade.
\Cref{tab:bpe_ctc_search_space,tab:bpe_transducer_search_space} show LibriSpeech lexicon-constrained \gls*{BPE} \gls*{CTC} and \gls*{BPE} transducer search with a word-level Transformer \gls*{LM}.
Search errors count segments that were transcribed incorrectly although the ground-truth transcription has a better score than the recognized one, i.e., the ground-truth was erroneously pruned away during search.
The tables also include percentile statistics of the within-word hypothesis counts at each search step.
\begin{figure}
    \includegraphics{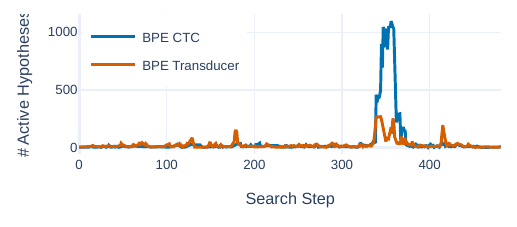}
    \caption{Illustrative number of active within-word hypotheses at each search step for a selected LibriSpeech dev-other utterance with \gls*{BPE} \gls*{CTC}/transducer using lexicon-constrained search including a word-level 4-gram \gls*{LM}.}
    \label{fig:active_hyps_over_time}
\end{figure}
The p50 value shows that the median active hypothesis set is small while the p90 and p99 statistics show that in some rare difficult regions the signal causes locally increased hypothesis counts.
This effect is visualized in \Cref{fig:active_hyps_over_time} which shows the hypothesis count by timestep for a selected utterance with a \gls*{CTC} and transducer model.
The search space stays small for most search steps and exhibits one large temporary peak between steps 300 and 400.

\Cref{tab:bpe_transducer_search_space_loquacious} reports the corresponding analysis for Loquacious with \gls*{BPE} transducer decoding and a word 4-gram \gls*{LM}.
Overall, the hypothesis counts are lower on Loquacious.
This effect can be explained by the fact that the experiments on Loquacious only use a 4-gram \gls*{LM} and a much lower \gls*{LM} scale (see \Cref{tab:tuned_parameters}).
Since the 4-gram only relies on a three-word history, the recombination step can recombine many more hypotheses and with the low \gls*{LM} scale, score jumps from the \gls*{LM} score contributions are much smaller and need a lower score threshold to cover for them.
Still, the same non-uniform behavior can be observed on this dataset: the median active set remains small, while p99 reflects rare high-complexity regions and is comparatively much higher.

To test more difficult acoustic conditions, we collected similar statistics for a phoneme \gls*{CTC} model on the in-house Spanish 8kHz telephony test set, as shown in \Cref{tab:phoneme_ctc_search_space}.
This task has a much higher \gls*{WER} and a low hypothesis count in the tens is not nearly enough to reach full performance in this case.
Even at the median, several hundred hypotheses are required and the p90 and p99 values go into the thousands in order to reach optimal performance.

These results confirm that small fixed beams are insufficient when model uncertainty is higher and when external \gls*{LM} contribution makes up a large part of the score.
Dynamic pruning adapts to the model uncertainty and \gls*{LM} score contributions and keeps the hypothesis counts comparatively low in the median.

\section{Search Efficiency}
\label{sec:efficiency}
Search efficiency is tied to the size of the active search space.
We compare fixed max-beam pruning with dynamic score-threshold pruning while preserving the \gls*{WER}.
For each max-beam size $B_\mathrm{ASR}$, the fixed baseline uses only max-beam pruning while the dynamic variant additionally performs score-threshold pruning with the tightest possible thresholds that do not degrade the \gls*{WER}.
We report the relative search \gls*{RTF} reduction when switching from fixed to dynamic pruning in percent:
\begin{equation}
    \mathrm{RTF\ reduction} \left[\% \right] =
    100 \cdot \left( 1 - \frac{\mathrm{RTF}_\mathrm{dynamic}}{\mathrm{RTF}_\mathrm{fixed}} \right).
\end{equation}
\begin{figure}
    \includegraphics{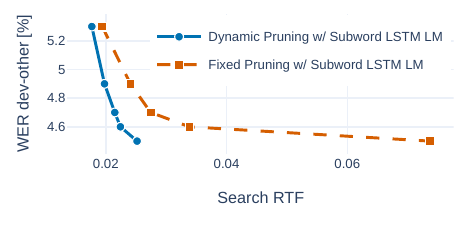}
    \caption{Search RTF vs. WER for fixed and dynamic pruning on LibriSpeech dev-other with \gls*{BPE} transducer + subword-level \gls*{LSTM} \gls*{LM}.}
    \label{fig:score_vs_beam_pruning_librispeech_subword}
\end{figure}
\begin{figure}
    \includegraphics{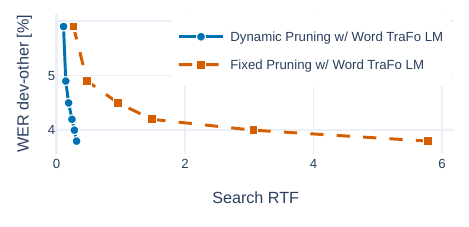}
    \caption{Search RTF vs. WER for fixed and dynamic pruning on LibriSpeech dev-other with \gls*{BPE} transducer + word-level Transformer \gls*{LM}.}
    \label{fig:score_vs_beam_pruning_librispeech_word}
\end{figure}
\begin{figure}
    \includegraphics{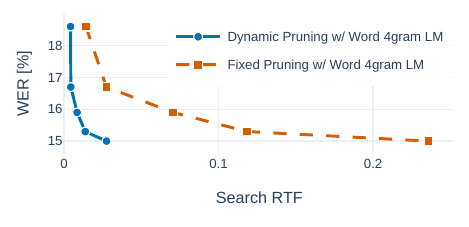}
    \caption{Search RTF vs. WER for fixed and dynamic pruning on Spanish telephony with phoneme \gls*{CTC} + word-level 4-gram \gls*{LM}.}
    \label{fig:score_vs_beam_pruning_spanish}
\end{figure}
\Cref{fig:score_vs_beam_pruning_librispeech_subword,fig:score_vs_beam_pruning_librispeech_word} contrast score-based pruning with pure max-beam pruning for LibriSpeech \gls*{BPE} transducer search with subword and word \glspl*{LM}.
In both cases, dynamic pruning achieves the same \glspl*{WER} at much faster inference times, especially at higher beam sizes.
With dynamic pruning, the \gls*{RTF} only increases marginally while fixed beam pruning quickly becomes very expensive.
For the smaller \gls*{LSTM} \gls*{LM} the search times are faster than with the larger Transformer \gls*{LM} but the relative difference between fixed and dynamic pruning is still clear.
\Cref{fig:score_vs_beam_pruning_spanish} shows the same effect on Spanish telephony with a 4-gram \gls*{LM}.

\Cref{tab:dyn_pruning_transducer_lstm_librispeech,tab:dyn_pruning_transducer_word_trafo_librispeech} provide detailed comparisons over the max beam size $B_\mathrm{ASR}$ for the subword-level \gls*{LSTM} \gls*{LM} and word-level Transformer \gls*{LM} setups on LibriSpeech, including the relative RTF reduction.
\begin{table}
\centering
\caption{WER and Search RTF breakdown for fixed and dynamic pruning with different max beam sizes $B_\mathrm{ASR}$ on LibriSpeech dev-other with BPE Transducer + subword-level LSTM LM.}
\label{tab:dyn_pruning_transducer_lstm_librispeech}
\begin{tabular}{|r|r|r|r|r|} 
\hline
\multicolumn{1}{|l|}{\multirow{2}{*}{$B_\mathrm{ASR}$}} & \multicolumn{1}{l|}{\multirow{2}{*}{WER [\%]}} & \multicolumn{2}{l|}{Search RTF $\cdot 10^3$}                           & \multicolumn{1}{l|}{\multirow{2}{*}{RTF reduction [\%]}}  \\ 
\cline{3-4}
\multicolumn{1}{|l|}{}                                & \multicolumn{1}{l|}{}                          & \multicolumn{1}{l|}{Fixed} & \multicolumn{1}{l|}{Dynamic} & \multicolumn{1}{l|}{}                                     \\ 
\hhline{|=====|}
4                                                     & 5.3                                            & 19                      & 18                        & 5.3                                                      \\ 
\hline
16                                                    & 4.7                                            & 27                      & 21                        & 22.2                                                      \\ 
\hline
64                                                    & 4.6                                            & 49                      & 24                        & 51.0                                                      \\ 
\hline
128                                                   & 4.5                                            & 74                      & 25                        & 66.2                                                      \\
\hline
\end{tabular}
\end{table}

\begin{table}
\centering
\caption{WER and Search RTF breakdown for fixed and dynamic pruning with max beam sizes $B_\mathrm{ASR}$ on LibriSpeech dev-other with BPE Transducer + word-level Transformer LM.}
\label{tab:dyn_pruning_transducer_word_trafo_librispeech}
\begin{tabular}{|r|r|r|r|r|} 
\hline
\multicolumn{1}{|c|}{\multirow{2}{*}{$B_\mathrm{ASR}$}} & \multicolumn{1}{c|}{\multirow{2}{*}{WER [\%]}} & \multicolumn{2}{c|}{Search RTF $\cdot 10^3$}                           & \multicolumn{1}{c|}{\multirow{2}{*}{RTF reduction [\%]}}  \\ 
\cline{3-4}
\multicolumn{1}{|c|}{}                                & \multicolumn{1}{c|}{}                          & \multicolumn{1}{c|}{Fixed} & \multicolumn{1}{c|}{Dynamic} & \multicolumn{1}{c|}{}                                     \\ 
\hhline{|=====|}
4                                                     & 10.4                                           & 72                      & 60                        & 16.7                                                      \\ 
\hline
16                                                    & 5.9                                            & 261                      & 107                        & 59.0                                                      \\ 
\hline
64                                                    & 4.5                                            & 955                      & 186                        & 80.5                                                      \\ 
\hline
256                                                   & 4.0                                            & 3063                      & 276                        & 91.0                                                      \\
\hline
1024                                                  & 3.8                                            & 5781                      & 310                        & 94.6                                                      \\
\hline
\end{tabular}
\end{table}

\begin{table*}
\centering
\caption{\gls*{RTF} reduction through dynamic pruning for different setups. Measured on dev-other for LibriSpeech, dev.all for Loquacious and an in-house test-set for Spanish. Avg. Hyps counts active subword-level hypotheses after pruning.}
\label{tab:dyn_pruning_speedup}
\begin{tabular}{|c|c|c|r|r|r|r|r|r|} 
\hline
\multicolumn{1}{|c|}{\multirow{2}{*}{Corpus}} & \multicolumn{1}{c|}{\multirow{2}{*}{Model}} & \multicolumn{1}{c|}{\multirow{2}{*}{LM}} & \multicolumn{1}{c|}{\multirow{2}{*}{WER [\%]~}} & \multicolumn{2}{c|}{Fixed Pruning}                               & \multicolumn{2}{c|}{Dynamic Pruning}                             & \multicolumn{1}{c|}{RTF}             \\ 
\cline{5-8}
\multicolumn{1}{|c|}{}                        & \multicolumn{1}{c|}{}                       & \multicolumn{1}{c|}{}                    & \multicolumn{1}{c|}{}                           & \multicolumn{1}{c|}{Avg. Hyps} & \multicolumn{1}{c|}{RTF $\cdot 10^3$} & \multicolumn{1}{c|}{Avg. Hyps} & \multicolumn{1}{c|}{RTF $\cdot 10^3$} & \multicolumn{1}{c|}{red. [\%]}  \\ 
\hhline{|=========|}
\multirow{10}{*}{LibriSpeech}                 & \multirow{4}{*}{BPE CTC}                    & -                                        & 7.0                                             & 1                            & 0.62                           & 1.0                            & 0.62                           & 0.0                                  \\ 
\cline{3-9}
                                              &                                             & Word 4-gram                              & 5.2                                             & 256                          & 326                           & 13.5                           & 1.6                           & 99.5                                 \\ 
\cline{3-9}
                                              &                                             & Subword LSTM                             & 5.1                                             & 128                          & 140                           & 27.7                           & 22                           & 84.3                                 \\ 
\cline{3-9}
                                              &                                             & Word Trafo                               & 4.6                                             & 256                          & 1203                           & 61.2                           & 154                           & 87.2                                 \\ 
\cline{2-9}
                                              & \multirow{2}{*}{Phoneme CTC}                & Word 4-gram                              & 5.5                                             & 256                          & 128                           & 13.2                           & 1.5                           & 98.8                                 \\ 
\cline{3-9}
                                              &                                             & Word Trafo                               & 4.5                                             & 256                          & 1320                           & 60.0                           & 292                           & 77.9                                 \\ 
\cline{2-9}
                                              & \multirow{4}{*}{BPE Transducer}             & -                                        & 6.0                                             & 1                            & 7.6                           & 1.0                            & 7.6                           & 0.0                                  \\ 
\cline{3-9}
                                              &                                             & Word 4-gram                              & 4.8                                             & 128                          & 97                           & 12.7                           & 13                           & 86.6                                 \\ 
\cline{3-9}
                                              &                                             & Subword LSTM                             & 4.5                                             & 128                          & 74                           & 13.8                           & 25                           & 66.2                                 \\ 
\cline{3-9}
                                              &                                             & Word Trafo                               & 3.8                                             & 1024                         & 5781                           & 150.6                          & 310                           & 94.6                                \\ 
\hhline{|=========|}
\multirow{5}{*}{Loquacious}                   & \multirow{2}{*}{BPE CTC}                    & -                                        & 11.7                                            & 1                            & 0.62                           & 1.0                            & 0.62                           & 0.0                                  \\ 
\cline{3-9}
                                              &                                             & Word 4-gram                              & 9.5                                             & 64                           & 91                           & 4.2                            & 0.89                           & 99.0                                 \\ 
\cline{2-9}
                                              & \multirow{2}{*}{BPE Transducer}             & -                                        & 9.0                                             & 1                            & 7.7                           & 1.0                            & 7.7                           & 0.0                                  \\ 
\cline{3-9}
                                              &                                             & Word 4-gram                              & 8.0                                             & 64                           & 121                           & 2.9                            & 13                           & 89.3                                 \\ 
\cline{2-9}
                                              & Phoneme Transducer                          & Word 4-gram                              & 8.4                                             & 64                           & 70                           & 4.6                            & 13                           & 81.4                                 \\ 
\hhline{|=========|}
Spanish In-House                              & Phoneme CTC                                 & Word 4-gram                              & 14.9                                            & 4096                         & 1860                           & 932.2                          & 90                           & 95.2                                 \\
\hline
\end{tabular}
\end{table*}

\begin{table*}
\centering
\caption{Tuned decoding parameters for dynamic pruning setups.}
\label{tab:tuned_parameters}
\begin{tabular}{|c|c|c|r|r|r|r|r|r|} 
\hline
Corpus                        & Model                           & LM           & \multicolumn{1}{c|}{LM Scale} & \multicolumn{1}{l|}{ILM / Prior Scale} & \multicolumn{1}{l|}{$B_\text{ASR}$} & \multicolumn{1}{l|}{$B_\text{LM}$} & \multicolumn{1}{l|}{$\Delta_\text{ASR}$} & \multicolumn{1}{l|}{$\Delta_\text{LM}$}  \\ 
\hhline{|=========|}
\multirow{10}{*}{LibriSpeech} & \multirow{4}{*}{BPE CTC}        & -            & -                             & -                                      & 1                                   & -                                  & 0.0                                      & -                                        \\ 
\cline{3-9}
                              &                                 & Word 4-gram  & 0.6                           & 0.2                                    & 256                                 & 16                                 & 12.0                                     & 6.0                                      \\ 
\cline{3-9}
                              &                                 & Subword LSTM & 0.8                           & 0.2                                    & 1024                                & 128                                & 14.0                                     & 12.0                                     \\ 
\cline{3-9}
                              &                                 & Word Trafo   & 0.8                           & 0.2                                    & 256                                 & 16                                 & 16.0                                     & 8.0                                      \\ 
\cline{2-9}
                              & \multirow{2}{*}{Phoneme CTC}    & Word 4-gram  & 0.6                           & 0.2                                    & 256                                 & 8                                  & 12.0                                     & 6.0                                      \\ 
\cline{3-9}
                              &                                 & Word Trafo   & 0.8                           & 0.2                                    & 256                                 & 16                                 & 16.0                                     & 8.0                                      \\ 
\cline{2-9}
                              & \multirow{4}{*}{BPE Transducer} & -            & -                             & -                                      & 1                                   & -                                  & 0.0                                      & -                                        \\ 
\cline{3-9}
                              &                                 & Word 4-gram  & 0.6                           & 0.2                                    & 128                                 & 4                                  & 12.0                                     & 4.8                                      \\ 
\cline{3-9}
                              &                                 & Subword LSTM & 0.8                           & 0.2                                    & 64                                  & 32                                 & 10.0                                     & 8.0                                      \\ 
\cline{3-9}
                              &                                 & Word Trafo   & 0.8                           & 0.2                                    & 1024                                & 64                                 & 16.0                                     & 12.8                                     \\ 
\hhline{|=========|}
\multirow{5}{*}{Loquacious}   & \multirow{2}{*}{BPE CTC}        & -            & -                             & -                                      & 1                                   & -                                  & 0.0                                      & -                                        \\ 
\cline{3-9}
                              &                                 & Word 4-gram  & 0.3                           & 0.2                                    & 64                                  & 16                                 & 6.0                                      & 3.0                                      \\ 
\cline{2-9}
                              & \multirow{2}{*}{BPE Transducer} & -            & -                             & -                                      & 1                                   & -                                  & 0.0                                      & -                                        \\ 
\cline{3-9}
                              &                                 & Word 4-gram  & 0.3                           & 0.2                                    & 64                                  & 16                                 & 6.0                                      & 3.0                                      \\ 
\cline{2-9}
                              & Phoneme Transducer              & Word 4-gram  & 0.3                           & 0.2                                    & 64                                  & 16                                 & 6.0                                      & 3.0                                      \\ 
\hhline{|=========|}
Spanish In-House              & Phoneme CTC                     & Word 4-gram  & 0.5                           & 0.3                                    & 4096                                & 16                                 & 10.0                                     & 5.0                                      \\
\hline
\end{tabular}
\end{table*}

In \Cref{tab:dyn_pruning_speedup} we summarize the reduction in average hypothesis count and search \gls*{RTF} for different \gls*{ASR} models and external \glspl*{LM} on all three datasets.
When no external \gls*{LM} is used, greedy search already gives optimal results and dynamic pruning provides no advantage in this case.
For all experiments with external \gls*{LM}, however, dynamic pruning provides significant efficiency improvements to reach the same \gls*{WER}.
The average hypothesis count and search \gls*{RTF} are often reduced by around an order of magnitude.
The relative \gls*{RTF} reduction is largest for \gls*{CTC} and 4-gram \gls*{LM} setups because their search cost is dominated by hypothesis expansion and CPU-side bookkeeping, while transducer prediction/joiner networks and neural \glspl*{LM} use batched GPU computations which do not scale linearly with the active set.

\Cref{tab:tuned_parameters} lists tuned dynamic-pruning parameters.
The \gls*{LM} scale weights external-\gls*{LM} scores; the \gls*{ILM}/prior scale controls transducer \gls*{ILM} or \gls*{CTC} prior correction.
$B_\mathrm{ASR}$/$B_\mathrm{LM}$ are maximum beam sizes and $\Delta_\mathrm{ASR}$/$\Delta_\mathrm{LM}$ are score thresholds pruning at the \gls*{ASR} and \gls*{LM} stages respectively.
These parameters define the pruning settings used in \Cref{tab:dyn_pruning_speedup}.

\section{Conclusions}
\label{sec:conclusions}
We presented a systematic comparison of search strategies for time-synchronous \gls*{BPE}- and phoneme-based \gls*{CTC} and monotonic neural transducer models in \gls*{E2E} \gls*{ASR}.
Comparisons of \gls*{WER} and \gls*{RTF} show that dynamic score-based pruning and external \gls*{LM} combination remain relevant for decoding with modern \gls*{E2E} \gls*{ASR} models.
We have seen that typically, these models operate with a low active hypothesis count for the majority of search steps but spike to hundreds or thousands of hypotheses in rare difficult regions.
This behavior is verified on the public LibriSpeech task as well as the more heterogeneous public Loquacious task and a more difficult in-house telephony dataset.

The fixed-vs-dynamic pruning experiments show that dynamic pruning can preserve \gls*{WER} while reducing search \gls*{RTF} by adapting the active hypothesis set to local uncertainty, especially with external \glspl*{LM} where low fixed-size beams are insufficient for optimal performance.
We provide tuned decoding parameters and open source implementations for reproducibility.

This study is limited to time-synchronous \gls*{CTC} and transducer search; label-synchronous and streaming decoding require some fundamentally different analysis and remain future work.

Overall, our results suggest that external \gls*{LM} integration and dynamic pruning are underutilized in modern \gls*{E2E} \gls*{ASR} applications.

\begin{center}\textbf{Acknowledgements}\end{center}
\label{sec:acknowledgements}
\ifanonymous
    Anonymized for double-blind review.
\else
    This work was partially supported by the project RESCALE within the program \textit{AI Lighthouse Projects for the Environment, Climate, Nature and Resources} funded by the Federal Ministry for the Environment, Nature Conservation, Nuclear Safety and Consumer Protection (BMUV), funding ID: 67KI32006A.

\fi

\clearpage

\bibliographystyle{IEEEbib}
\bibliography{refs}

\end{document}